\documentclass[11pt,onecolumn]{article}
\usepackage[utf8]{inputenc}
\usepackage{amsmath,amsfonts,amssymb,graphicx,authblk,placeins,cite}
\usepackage{comment}
\usepackage{color,soul}
\usepackage{hyphenat}
\usepackage[top=50pt,bottom=50pt,left=70pt,right=70pt]{geometry}
\usepackage{sectsty}
\usepackage[font=sf]{caption}
\usepackage{hyperref}
\usepackage{graphicx}
\usepackage{subfig}
\begin{document}
\allsectionsfont{\sffamily}
\pagenumbering{gobble}


\title{\sffamily \Huge Word class representations spontaneously emerge in a deep neural network trained on next word prediction
 \vspace{1 cm}}


\author[1]{\sffamily Kishore Surendra}
\author[2,3]{\sffamily Achim Schilling}
\author[3,4]{\sffamily Paul Stoewer}
\author[4]{\sffamily Andreas Maier}
\author[2,3,5]{\sffamily Patrick Krauss}


\affil[1]{\small University Hospital Hamburg, Germany}

\affil[2]{\small{Neuroscience Lab, University Hospital Erlangen, Germany}}

\affil[3]{\small{Cognitive Computational Neuroscience Group, University Erlangen-Nuremberg, Germany}}

\affil[4]{\small{Pattern Recognition Lab, University Erlangen-Nuremberg, Germany}}

\affil[5]{\small Linguistics Lab, University Erlangen-Nuremberg, Germany}

\maketitle


\begin{abstract}{\sffamily \noindent
How do humans learn language, and can the first language be learned at all? These fundamental questions are still hotly debated. In contemporary linguistics, there are two major schools of thought that give completely opposite answers. According to Chomsky's theory of universal grammar, language cannot be learned because children are not exposed to sufficient data in their linguistic environment. In contrast, usage-based models of language assume a profound relationship between language structure and language use. In particular, contextual mental processing and mental representations are assumed to have the cognitive capacity to capture the complexity of actual language use at all levels. The prime example is syntax, i.e., the rules by which words are assembled into larger units such as sentences. Typically, syntactic rules are expressed as sequences of word classes. However, it remains unclear whether word classes are innate, as implied by universal grammar, or whether they emerge during language acquisition, as suggested by usage-based approaches. Here, we address this issue from a machine learning and natural language processing perspective. In particular, we trained an artificial deep neural network on predicting the next word, provided sequences of consecutive words as input. Subsequently, we analyzed the emerging activation patterns in the hidden layers of the neural network. Strikingly, we find that the internal representations of nine-word input sequences cluster according to the word class of the tenth word to be predicted as output, even though the neural network did not receive any explicit information about syntactic rules or word classes during training. This surprising result suggests, that also in the human brain, abstract representational categories such as word classes may naturally emerge as a consequence of predictive coding and processing during language acquisition.
}
\end{abstract}




\section*{Introduction}

The question of how humans come to language is one of the oldest scientific problems \cite{zimmer1986so}. According to the Greek historian Herodotus, already 2500 years ago the Egyptian pharaoh Psamtik sought to discover the origin of language. Therefore, he conducted an experiment with two children which he gave as newborn babies to a shepherd who should feed and care for them, but had the instruction not to speak to them. Psametik hypothesized that the infants' first word would be uttered in the root language of all people. Consequently, as one of the children cried \emph{'bekos'} which was the sound of the Phrygian word for "bread", Psamtik concluded that Phrygian was the root language of all humans because that \cite{rawlinson1861history}. Obviously, the assumption behind this cruel language deprivation experiment was that humans are born with innate words and their meanings, and that this root language is somehow 'over-ruled' during individual development and first language learning.

Nowadays, it is of course clear that words and meanings are not innate but rather learned during language acquisition \cite{goodluck1991language}, and that there is no causal relation between the signifier (sound pattern) and the signified (meaning) \cite{de2011course}. However, it is still highly debated to what extent language capacities are innate or must be learned.

According to Chomsky's theory of universal grammar, humans have an innate, genetically determined language faculty that e.g. distinguishes between different word classes such as nouns and verbs making it easier and faster for children to learn to speak \cite{chomsky2012nature, chomsky2014aspects, yang2017growth}. In contrast, in cognitive linguistics and usage-based approaches, a profound relationship between language structure and language use is assumed \cite{goldberg1995constructions, goldberg2003constructions, tomasello2005constructing, langacker2008cognitive}. In particular, contextual mental processing and mental representations are assumed to have the cognitive capacity to capture the complexity of actual language use at all levels \cite{bybee1994evolution, hopper2001frequency, bybee2013usage, diessel2019usage, goldberg2019explain, schmid2020dynamics}. According to Diessel, grammar is a "dynamic system of emergent structures" and it needs to be explained "how linguistic structures evolve" during language acquisition \cite{diessel201514}.

Predictive coding and processing are thought to be canonical computations of the human brain \cite{kilner2007predictive,bastos2012canonical,keller2018predictive,schilling2022predictive}, in particular during speech and language processing which involves the prediction of which words come next \cite{garibyan2022neural}. In previous studies, we already demonstrated that efficient successor representations to form cognitive maps of space and language can be learned by artificial neural networks \cite{stoewer2022Aneural,stoewer2022Bneural}. In particular, we demonstrated how a neural network model can infer the underlying word classes of a simplified artificial language just by observing sequences of words, i.e. sentences, and without any prior knowledge about actual word classes or grammar. The emerging representations share important properties with network-like cognitive maps, enabling e.g. navigation in arbitrary abstract and conceptual spaces, and thereby broadly supporting domain-general cognition, as proposed by Bellmund et al. \cite{bellmund2018navigating}. 

In this follow-up study, we further address the question if abstract linguistic categories and structures can be learned from experienced language alone in a more complex and naturalistic linguistic task, i.e. word prediction in a natural language scenario. In particular, we trained an artificial deep neural network to predict the next word (successor) in a novel given the nine consecutive predecessor words as input. Subsequently, we analyzed the emerging activation patterns in the hidden layers of the neural network. Strikingly, we find that the internal representations of nine-word input sequences cluster according to the word class of the tenth word to be predicted as output, even though the neural network did not receive any explicit information about syntactic rules or word classes during training. 
This surprising result suggests, that also in the human brain, abstract representational categories such as word classes may naturally emerge as a consequence of predictive coding and processing of language input. Based on these findings we hypothesize that during language acquisition -- which at least partly corresponds to learn to predict which word or utterance comes next --, word classes spontaneously emerge as clusters of successor representations of perceived utterances. We conclude that word classes need not to be innate to enable efficient language acquisition as suggested by universal grammar.

\section*{Methods}

\subsection*{Data pre processing}

The German novel \emph{Gut gegen Nordwind} by Daniel Glattauer (\copyright \; \emph{Deuticke im Paul Zsolnay Verlag}, Wien 2006, published by \emph{Deuticke Verlag} served as natural language text data for training and testing our model. The complete text data consists of a total number of 40460 tokens and 6117 types. Prior to further analysis, punctuation and special characters have been removed from the text corpus. Furthermore, repetitive words and extra white spaces have been removed, and all numbers have been replaced by a single word (cf. table \ref{tab1}). All words are converted to lower case to maintain uniformity, so that the same word occurring in a different case is considered as two tokens of the same type, instead of two different types. All words have been encoded as 384-dimensional word vectors using the word2vec embedding function from the python library \emph{spaCy} \cite{explosion2017spacy}. Sequences of nine consecutive word vectors served as input, while one (the tenth) or two (tenth and eleventh) word vectors served as corresponding output. Finally, the all word vector sequences were split into a training (chapters 1 to 7 of the novel) and a test data set (chapters 8 and 9 of the novel). 

\begin{table}[!htb]
\centering
    \begin{tabular}{ |p{10cm}|p{4cm}| }
    \hline
    \textbf{Character/Word} & \textbf{Operation} \\
    \hline
    Repititive words:RE:,AW:,Eine,Zwei,..,Stunden,Sekunden,&\\ Stunden...,sp{\"a}ter,Am n{\"a}chsten,Kein Betreff,Betreff & Remove completely\\
    \hline
    Punctuation and other characters: .,',\",?,\%,\&,',',! & Remove completely\\
    \hline
    Numbers: 18,1,500,... & Replace with 'nummer'\\
    \hline
    Extra whitespaces & Replace with single space\\
    \hline
    E-mail & Replace with 'email'\\
    \hline
    \end{tabular}
    \caption{\textbf{Data cleaning.} Words, characters and their replacements during data cleaning.}
    \label{tab1}
\end{table}
\FloatBarrier

\subsection*{Neural network architecture and training procedure}

For the task at hand, i.e. to predict the tenth word (or the tenth and the eleventh word), given a prior sequence of nine words occurring in the corpus, recurrent neural networks (RNNs) are perfectly suited. Here, we implemented a neural network consisting of four bi-directional LSTM (long short-term memories) layers (with 128, 128, 64, and 64 neurons) followed by a flatten layer, and a dense output layer (384 neurons). The input consisted of sequences of nine 384-dimensional word embedding vectors generated as described above. The expected output is a single (or a sequence of two) 384-dimensional word embedding vectors. Weights were initialized using the Glorot uniform initialization, which is Keras's default initializer. As optimizer, we used Adam with a learning rate of $0.001$ and as loss function we used mean-squared error. Training was performed for 100 epochs.

\subsection*{Word classes}

Word classes were analysed by applying \emph{part-of-speech (POS) tagging} \cite{ratnaparkhi1996maximum, marquez1998part, jurafsky2014speech} as implemented in the python library \emph{spaCy} \cite{explosion2017spacy}.
The used POS tags comprised the following 13 default word classes: ’NUM’, ’VERB’, ’ADJ’, ’X’, 'PART’, ’NOUN’, ’SCONJ’,’ADP’, ’DET’, ’PRON’, ’CONJ’, ’AUX’ and ’ADV’. Their exact definitions can be found in \cite{petrov2011universal}.
Note that, during training the neural networks, we did not provide any information about word classes as input.

\subsection*{Multi-dimensional scaling}

A frequently used method to generate low-dimensional embeddings of high-dimensional data is t-distributed stochastic neighbor embedding (t-SNE) \cite{van2008visualizing}. However, in t-SNE the resulting low-dimensional projections can be highly dependent on the detailed parameter settings \cite{wattenberg2016use}, sensitive to noise, and may not preserve, but rather often scramble the global structure in data \cite{vallejos2019exploring, moon2019visualizing}.
In contrast to that, multi-Dimensional-Scaling (MDS) \cite{torgerson1952multidimensional, kruskal1964nonmetric,kruskal1978multidimensional,cox2008multidimensional} is an efficient embedding technique to visualize high-dimensional point clouds by projecting them onto a 2-dimensional plane. Furthermore, MDS has the decisive advantage that it is parameter-free and all mutual distances of the points are preserved, thereby conserving both the global and local structure of the underlying data. 

When interpreting patterns as points in high-dimensional space and dissimilarities between patterns as distances between corresponding points, MDS is an elegant method to visualize high-dimensional data. By color-coding each projected data point of a data set according to its label, the representation of the data can be visualized as a set of point clusters. For instance, MDS has already been applied to visualize for instance word class distributions of different linguistic corpora \cite{schilling2021analysis}, hidden layer representations (embeddings) of artificial neural networks \cite{schilling2021quantifying,krauss2021analysis}, structure and dynamics of highly recurrent neural networks \cite{krauss2019analysis, krauss2019recurrence, krauss2019weight, metzner2023quantifying}, or brain activity patterns assessed during e.g. pure tone or speech perception \cite{krauss2018statistical,schilling2021analysis}, or even during sleep \cite{krauss2018analysis,traxdorf2019microstructure,metzner2022classification,metzner2023extracting}. 
In all these cases the apparent compactness and mutual overlap of the point clusters permits a qualitative assessment of how well the different classes separate.

\subsection*{Generalized Discrimination Value (GDV)}

We used the GDV to calculate cluster separability as published and explained in detail in \cite{schilling2021quantifying}. Briefly, we consider $N$ points $\mathbf{x_{n=1..N}}=(x_{n,1},\cdots,x_{n,D})$, distributed within $D$-dimensional space. A label $l_n$ assigns each point to one of $L$ distinct classes $C_{l=1..L}$. In order to become invariant against scaling and translation, each dimension is separately z-scored and, for later convenience, multiplied with $\frac{1}{2}$:
\begin{align}
s_{n,d}=\frac{1}{2}\cdot\frac{x_{n,d}-\mu_d}{\sigma_d}.
\end{align}
Here, $\mu_d=\frac{1}{N}\sum_{n=1}^{N}x_{n,d}\;$ denotes the mean, and $\sigma_d=\sqrt{\frac{1}{N}\sum_{n=1}^{N}(x_{n,d}-\mu_d)^2}$ the standard deviation of dimension $d$.
Based on the re-scaled data points $\mathbf{s_n}=(s_{n,1},\cdots,s_{n,D})$, we calculate the {\em mean intra-class distances} for each class $C_l$ 
\begin{align}
\bar{d}(C_l)=\frac{2}{N_l (N_l\!-\!1)}\sum_{i=1}^{N_l-1}\sum_{j=i+1}^{N_l}{d(\textbf{s}_{i}^{(l)},\textbf{s}_{j}^{(l)})},
\end{align}
and the {\em mean inter-class distances} for each pair of classes $C_l$ and $C_m$
\begin{align}
\bar{d}(C_l,C_m)=\frac{1}{N_l  N_m}\sum_{i=1}^{N_l}\sum_{j=1}^{N_m}{d(\textbf{s}_{i}^{(l)},\textbf{s}_{j}^{(m)})}.
\end{align}
Here, $N_k$ is the number of points in class $k$, and $\textbf{s}_{i}^{(k)}$ is the $i^{th}$ point of class $k$.
The quantity $d(\textbf{a},\textbf{b})$ is the euclidean distance between $\textbf{a}$ and $\textbf{b}$. Finally, the Generalized Discrimination Value (GDV) is calculated from the mean intra-class and inter-class distances  as follows:
\begin{align}
\mbox{GDV}=\frac{1}{\sqrt{D}}\left[\frac{1}{L}\sum_{l=1}^L{\bar{d}(C_l)}\;-\;\frac{2}{L(L\!-\!1)}\sum_{l=1}^{L-1}\sum_{m=l+1}^{L}\bar{d}(C_l,C_m)\right]
 \label{GDVEq}
\end{align}

\noindent whereas the factor $\frac{1}{\sqrt{D}}$ is introduced for dimensionality invariance of the GDV with $D$ as the number of dimensions.

\vspace{0.2cm}\noindent Note that the GDV is invariant with respect to a global scaling or shifting of the data (due to the z-scoring), and also invariant with respect to a permutation of the components in the $N$-dimensional data vectors (because the euclidean distance measure has this symmetry). The GDV is zero for completely overlapping, non-separated clusters, and it becomes more negative as the separation increases. A GDV of -1 signifies already a very strong separation.

\subsection*{Code Implementation}

The models were coded in Python. The neural networks were designed using the Keras \cite{keras} and Keras-RL \cite{kerasrl} libraries. Mathematical operations were performed with numpy \cite{numpy} and scikit-learn \cite{scikit-learn} libraries.
Visualizations were realised with matplotlib \cite{matplot} and networkX \cite{networkX}. For natural language processing we used SpaCy \cite{explosion2017spacy}.


\section*{Results}

\subsection*{Next word prediction}

\begin{figure}[!htb]
  \centering
  \includegraphics[scale=0.5]{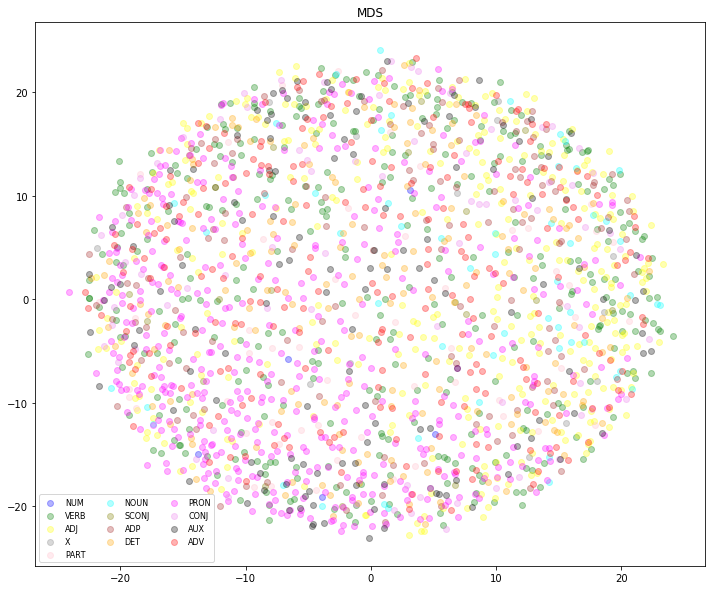}
  \label{fig1}
  \caption{Layer 1 results of neural network testing and projection onto a 2-dimensional plane using MDS, with color coding according to subsequent word class. The points are randomly distributed.}
\end{figure}

\begin{figure}[!htb]
  \centering
  \includegraphics[scale=0.5]{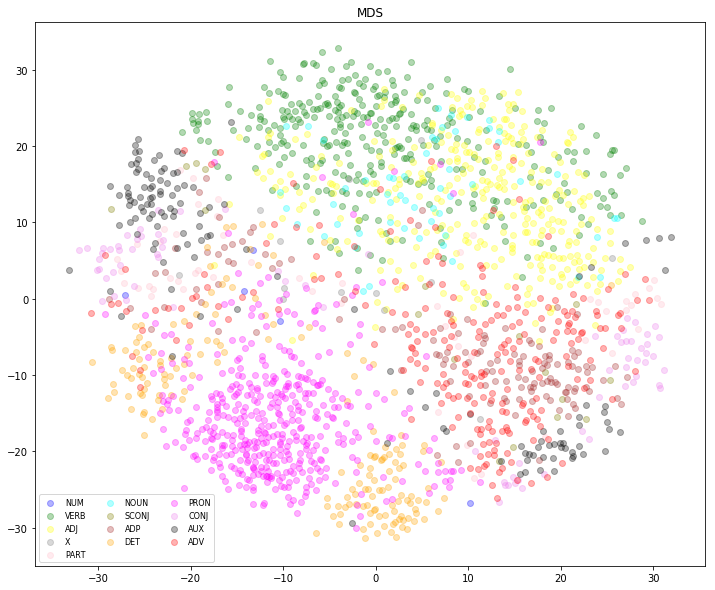}
  \label{fig2}
  \caption{Last layer results of neural network testing and projection onto a 2-dimensional plane using MDS, with color coding according to subsequent word class. The final layer shows strong clustering by word class, indicating that the neural network organizes internal representations based on the predicted subsequent word's class.}
\end{figure}

\begin{figure}[!htb]
  \centering
  \includegraphics[scale=0.8]{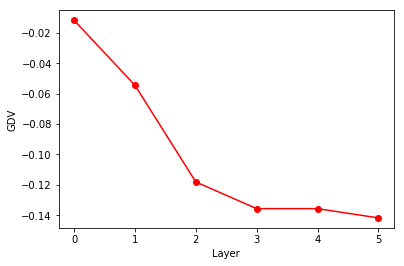}
  \label{fig3}
  \caption{GDV curve across layers of the neural network. The decline of the GDV indicates that the neural network has learned to cluster internal representations according to the subsequent word's class with increasingly strong clustering from input to output layer.}
\end{figure}

We trained a neural network on next word prediction using sequences of nine consecutive word vectors as input. The trained network was tested with sequences of nine words not used for training. The resulting neural activation of each layer was read out and the corresponding activation vectors were projected onto a 2-dimensional plane using MDS. All projected points were then color coded according to the word class of the subsequent word of the corresponding input sequence. Word classes were assessed after training using POS tagging and did not serve as input during training. While layer 1 shows a random distribution of the data points \ref{fig1}, we find a remarkably strong clustering according to world classes in the last layer of the neural network \ref{fig2}. This is also confirmed by the corresponding GDV curve across the layers \ref{fig3}. This means that the neural network organizes its internal representations of input word sequences according to the word class of the next word to be predicted.

\FloatBarrier
\newpage

\section*{Discussion}

The results of our study provide evidence that abstract linguistic categories, such as word classes, can emerge spontaneously in neural representations of linguistic input. This finding challenges the notion that the ability to recognize and categorize words by their grammatical function is innate and hardwired in the human brain, as proposed by Chomsky's theory of universal grammar.
Our results suggest that language acquisition involves, at least in part, the learning of predictive structures and categories based on statistical regularities in the input, rather than relying solely on innate linguistic knowledge. This is consistent with the view that language is a complex adaptive system shaped by both biological and environmental factors.
It is interesting to note that the clustering of input sequences by word class is evident only in the last layer of the neural network, suggesting that the network may gradually learn and refine more abstract and complex features of language as information flows through its layers. This finding is consistent with the hierarchical nature of language processing, in which higher-level representations build on lower-level representations.
One potential application of our findings is in natural language processing, where understanding the organization of neural representations of language input can help improve language modeling, machine translation, and other related tasks. In addition, our study provides a starting point for further investigations into the neural mechanisms underlying language acquisition and processing.
In conclusion, our study provides compelling evidence that neural networks can spontaneously learn to organize their internal representations of language input according to abstract linguistic categories such as word classes. Our results support the view that language acquisition is a complex and dynamic process that relies on both innate mechanisms and statistical learning from environmental input.

\section*{Acknowledgments}

We are grateful to the publisher \textit{Deuticke Verlag} for the permission to use the novel \textit{Gut gegen Nordwind} by Daniel Glattauer for the present and future studies.

This work was funded by the Deutsche Forschungsgemeinschaft (DFG, German Research Foundation): grants KR\,5148/2-1 (project number 436456810), KR\,5148/3-1 (project number 510395418) and GRK\,2839 (project number 468527017) to PK, and grant SCHI\,1482/3-1 (project number 451810794) to AS, and by the Emerging Talents Initiative (ETI) of the University Erlangen-Nuremberg (grant 2019/2-Phil-01 to PK).

\section*{Author contributions}
KS, AS, and PK performed computer simulations. PK designed the study. PK, AS and AM supervised the study. KS prepared the figures. All authors discussed the results and wrote the manuscript.

\section*{Competing interests}
The authors declare no competing financial interests.

\bibliographystyle{unsrt}
\bibliography{literature}

\begin{thebibliography}{10}

\bibitem{zimmer1986so}
Dieter~E Zimmer.
\newblock {\em So kommt der Mensch zur Sprache: {\"u}ber Spracherwerb,
  Sprachentstehung und Sprache \& Denken}, volume~16.
\newblock Heyne TB, 1986.

\bibitem{rawlinson1861history}
Henry~Creswicke Rawlinson, John~Gardner Wilkinson, et~al.
\newblock {\em The history of Herodotus}, volume~1.
\newblock 1861.

\bibitem{goodluck1991language}
Helen Goodluck.
\newblock {\em Language acquisition: A linguistic introduction.}
\newblock Basil Blackwell, 1991.

\bibitem{de2011course}
Ferdinand De~Saussure.
\newblock {\em Course in general linguistics}.
\newblock Columbia University Press, 2011.

\bibitem{chomsky2012nature}
Noam Chomsky.
\newblock On the nature, use and acquisition of language.
\newblock In {\em Language and Meaning in Cognitive Science}, pages 13--32.
  Routledge, 2012.

\bibitem{chomsky2014aspects}
Noam Chomsky.
\newblock {\em Aspects of the Theory of Syntax}, volume~11.
\newblock MIT press, 2014.

\bibitem{yang2017growth}
Charles Yang, Stephen Crain, Robert~C Berwick, Noam Chomsky, and Johan~J
  Bolhuis.
\newblock The growth of language: Universal grammar, experience, and principles
  of computation.
\newblock {\em Neuroscience \& Biobehavioral Reviews}, 81:103--119, 2017.

\bibitem{goldberg1995constructions}
Adele~E Goldberg.
\newblock {\em Constructions: A construction grammar approach to argument
  structure}.
\newblock University of Chicago Press, 1995.

\bibitem{goldberg2003constructions}
Adele~E Goldberg.
\newblock Constructions: A new theoretical approach to language.
\newblock {\em Trends in cognitive sciences}, 7(5):219--224, 2003.

\bibitem{tomasello2005constructing}
Michael Tomasello.
\newblock {\em Constructing a language: A usage-based theory of language
  acquisition}.
\newblock Harvard university press, 2005.

\bibitem{langacker2008cognitive}
Ronald~W Langacker.
\newblock Cognitive grammar.
\newblock {\em Basic Readings}, page~29, 2008.

\bibitem{bybee1994evolution}
Joan Bybee, Revere Perkins, William Pagliuca, et~al.
\newblock {\em The Evolution of Grammar: Tense, Aspect, and Modality in the
  Languages of the World}.
\newblock University of Chicago Press, 1994.

\bibitem{hopper2001frequency}
Paul~J Hopper and Joan~L Bybee.
\newblock Frequency and the emergence of linguistic structure.
\newblock {\em Frequency and the Emergence of Linguistic Structure}, pages
  1--502, 2001.

\bibitem{bybee2013usage}
Joan~L Bybee.
\newblock Usage-based theory and exemplar representations of constructions.
\newblock 2013.

\bibitem{diessel2019usage}
Holger Diessel, Ewa Dabrowska, and Dagmar Divjak.
\newblock Usage-based construction grammar.
\newblock {\em Cognitive linguistics}, 2:50--80, 2019.

\bibitem{goldberg2019explain}
Adele Goldberg and Adele~E Goldberg.
\newblock {\em Explain me this}.
\newblock Princeton University Press, 2019.

\bibitem{schmid2020dynamics}
Hans-J{\"o}rg Schmid.
\newblock {\em The dynamics of the linguistic system: Usage,
  conventionalization, and entrenchment}.
\newblock Oxford University Press, 2020.

\bibitem{diessel201514}
Holger Diessel.
\newblock 14. usage-based construction grammar.
\newblock In {\em Handbook of cognitive linguistics}, pages 296--322. De
  Gruyter Mouton, 2015.

\bibitem{kilner2007predictive}
James~M Kilner, Karl~J Friston, and Chris~D Frith.
\newblock Predictive coding: an account of the mirror neuron system.
\newblock {\em Cognitive processing}, 8(3):159--166, 2007.

\bibitem{bastos2012canonical}
Andre~M Bastos, W~Martin Usrey, Rick~A Adams, George~R Mangun, Pascal Fries,
  and Karl~J Friston.
\newblock Canonical microcircuits for predictive coding.
\newblock {\em Neuron}, 76(4):695--711, 2012.

\bibitem{keller2018predictive}
Georg~B Keller and Thomas~D Mrsic-Flogel.
\newblock Predictive processing: a canonical cortical computation.
\newblock {\em Neuron}, 100(2):424--435, 2018.

\bibitem{schilling2022predictive}
Achim Schilling, William Sedley, Richard Gerum, Claus Metzner, Konstantin
  Tziridis, Andreas Maier, Holger Schulze, Fan-Gang Zeng, Karl~J Friston, and
  Patrick Krauss.
\newblock Predictive coding and stochastic resonance: Towards a unified theory
  of auditory (phantom) perception.
\newblock {\em arXiv preprint arXiv:2204.03354}, 2022.

\bibitem{garibyan2022neural}
Armine Garibyan, Achim Schilling, Claudia Boehm, Alexandra Zankl, and Patrick
  Krauss.
\newblock Neural correlates of linguistic collocations during continuous speech
  perception.
\newblock {\em bioRxiv}, 2022.

\bibitem{stoewer2022Aneural}
Paul Stoewer, Christian Schlieker, Achim Schilling, Claus Metzner, Andreas
  Maier, and Patrick Krauss.
\newblock Neural network based successor representations to form cognitive maps
  of space and language.
\newblock {\em Scientific Reports}, 12(1):1--13, 2022.

\bibitem{stoewer2022Bneural}
Paul Stoewer, Achim Schilling, Andreas Maier, and Patrick Krauss.
\newblock Neural network based formation of cognitive maps of semantic spaces
  and the emergence of abstract concepts.
\newblock {\em arXiv preprint arXiv:2210.16062}, 2022.

\bibitem{bellmund2018navigating}
Jacob~LS Bellmund, Peter G{\"a}rdenfors, Edvard~I Moser, and Christian~F
  Doeller.
\newblock Navigating cognition: Spatial codes for human thinking.
\newblock {\em Science}, 362(6415), 2018.

\bibitem{explosion2017spacy}
AI~Explosion.
\newblock spacy-industrial-strength natural language processing in python.
\newblock {\em URL: https://spacy. io}, 2017.

\bibitem{petrov2011universal}
Slav Petrov, Dipanjan Das, and Ryan McDonald.
\newblock A universal part-of-speech tagset.
\newblock {\em arXiv preprint arXiv:1104.2086}, 2011.

\bibitem{van2008visualizing}
Laurens Van~der Maaten and Geoffrey Hinton.
\newblock Visualizing data using t-sne.
\newblock {\em Journal of machine learning research}, 9(11), 2008.

\bibitem{wattenberg2016use}
Martin Wattenberg, Fernanda Vi{\'e}gas, and Ian Johnson.
\newblock How to use t-sne effectively.
\newblock {\em Distill}, 1(10):e2, 2016.

\bibitem{vallejos2019exploring}
Catalina~A Vallejos.
\newblock Exploring a world of a thousand dimensions.
\newblock {\em Nature biotechnology}, 37(12):1423--1424, 2019.

\bibitem{moon2019visualizing}
Kevin~R Moon, David van Dijk, Zheng Wang, Scott Gigante, Daniel~B Burkhardt,
  William~S Chen, Kristina Yim, Antonia van~den Elzen, Matthew~J Hirn, Ronald~R
  Coifman, et~al.
\newblock Visualizing structure and transitions in high-dimensional biological
  data.
\newblock {\em Nature biotechnology}, 37(12):1482--1492, 2019.

\bibitem{torgerson1952multidimensional}
Warren~S Torgerson.
\newblock Multidimensional scaling: I. theory and method.
\newblock {\em Psychometrika}, 17(4):401--419, 1952.

\bibitem{kruskal1964nonmetric}
Joseph~B Kruskal.
\newblock Nonmetric multidimensional scaling: a numerical method.
\newblock {\em Psychometrika}, 29(2):115--129, 1964.

\bibitem{kruskal1978multidimensional}
Joseph~B Kruskal and Myron Wish.
\newblock {\em Multidimensional scaling}, volume~11.
\newblock Sage, 1978.

\bibitem{cox2008multidimensional}
Michael~AA Cox and Trevor~F Cox.
\newblock Multidimensional scaling.
\newblock In {\em Handbook of data visualization}, pages 315--347. Springer,
  2008.

\bibitem{schilling2021analysis}
Achim Schilling, Rosario Tomasello, Malte~R Henningsen-Schomers, Alexandra
  Zankl, Kishore Surendra, Martin Haller, Valerie Karl, Peter Uhrig, Andreas
  Maier, and Patrick Krauss.
\newblock Analysis of continuous neuronal activity evoked by natural speech
  with computational corpus linguistics methods.
\newblock {\em Language, Cognition and Neuroscience}, 36(2):167--186, 2021.

\bibitem{schilling2021quantifying}
Achim Schilling, Andreas Maier, Richard Gerum, Claus Metzner, and Patrick
  Krauss.
\newblock Quantifying the separability of data classes in neural networks.
\newblock {\em Neural Networks}, 139:278--293, 2021.

\bibitem{krauss2021analysis}
Patrick Krauss, Claus Metzner, Nidhi Joshi, Holger Schulze, Maximilian
  Traxdorf, Andreas Maier, and Achim Schilling.
\newblock Analysis and visualization of sleep stages based on deep neural
  networks.
\newblock {\em Neurobiology of sleep and circadian rhythms}, 10:100064, 2021.

\bibitem{krauss2019analysis}
Patrick Krauss, Alexandra Zankl, Achim Schilling, Holger Schulze, and Claus
  Metzner.
\newblock Analysis of structure and dynamics in three-neuron motifs.
\newblock {\em Frontiers in Computational Neuroscience}, 13:5, 2019.

\bibitem{krauss2019recurrence}
Patrick Krauss, Karin Prebeck, Achim Schilling, and Claus Metzner.
\newblock Recurrence resonance” in three-neuron motifs.
\newblock {\em Frontiers in computational neuroscience}, page~64, 2019.

\bibitem{krauss2019weight}
Patrick Krauss, Marc Schuster, Verena Dietrich, Achim Schilling, Holger
  Schulze, and Claus Metzner.
\newblock Weight statistics controls dynamics in recurrent neural networks.
\newblock {\em PloS one}, 14(4):e0214541, 2019.

\bibitem{metzner2023quantifying}
Claus Metzner, Marius~E Yamakou, Dennis Voelkl, Achim Schilling, and Patrick
  Krauss.
\newblock Quantifying and maximizing the information flux in recurrent neural
  networks.
\newblock {\em arXiv preprint arXiv:2301.12892}, 2023.

\bibitem{krauss2018statistical}
Patrick Krauss, Claus Metzner, Achim Schilling, Konstantin Tziridis, Maximilian
  Traxdorf, Andreas Wollbrink, Stefan Rampp, Christo Pantev, and Holger
  Schulze.
\newblock A statistical method for analyzing and comparing spatiotemporal
  cortical activation patterns.
\newblock {\em Scientific reports}, 8(1):1--9, 2018.

\bibitem{krauss2018analysis}
Patrick Krauss, Achim Schilling, Judith Bauer, Konstantin Tziridis, Claus
  Metzner, Holger Schulze, and Maximilian Traxdorf.
\newblock Analysis of multichannel eeg patterns during human sleep: a novel
  approach.
\newblock {\em Frontiers in human neuroscience}, 12:121, 2018.

\bibitem{traxdorf2019microstructure}
Maximilian Traxdorf, Patrick Krauss, Achim Schilling, Holger Schulze, and
  Konstantin Tziridis.
\newblock Microstructure of cortical activity during sleep reflects respiratory
  events and state of daytime vigilance.
\newblock {\em Somnologie}, 23(2):72--79, 2019.

\bibitem{metzner2022classification}
Claus Metzner, Achim Schilling, Maximilian Traxdorf, Konstantin Tziridis,
  Andreas Maier, Holger Schulze, and Patrick Krauss.
\newblock Classification at the accuracy limit: facing the problem of data
  ambiguity.
\newblock {\em Scientific Reports}, 12(1):22121, 2022.

\bibitem{metzner2023extracting}
Claus Metzner, Achim Schilling, Maximilian Traxdorf, Holger Schulze, Konstantin
  Tziridis, and Patrick Krauss.
\newblock Extracting continuous sleep depth from eeg data without machine
  learning.
\newblock {\em arXiv preprint arXiv:2301.06755}, 2023.

\bibitem{keras}
Fran\c{c}ois Chollet et~al.
\newblock Keras.
\newblock \url{https://keras.io}, 2015.
\newblock Last visited: \today.

\bibitem{kerasrl}
Matthias Plappert.
\newblock keras-rl.
\newblock \url{https://github.com/keras-rl/keras-rl}, 2016.
\newblock Last visited: \today.

\bibitem{numpy}
Charles~R. Harris, K.~Jarrod Millman, St{'{e}}fan~J. van~der Walt, Ralf
  Gommers, Pauli Virtanen, David Cournapeau, Eric Wieser, Julian Taylor,
  Sebastian Berg, Nathaniel~J. Smith, Robert Kern, Matti Picus, Stephan Hoyer,
  Marten~H. van Kerkwijk, Matthew Brett, Allan Haldane, Jaime~Fern{'{a}}ndez
  del R{'{\i}}o, Mark Wiebe, Pearu Peterson, Pierre G{'{e}}rard-Marchant, Kevin
  Sheppard, Tyler Reddy, Warren Weckesser, Hameer Abbasi, Christoph Gohlke, and
  Travis~E. Oliphant.
\newblock Array programming with {NumPy}.
\newblock {\em Nature}, 585(7825):357--362, September 2020.

\bibitem{scikit-learn}
F.~Pedregosa, G.~Varoquaux, A.~Gramfort, V.~Michel, B.~Thirion, O.~Grisel,
  M.~Blondel, P.~Prettenhofer, R.~Weiss, V.~Dubourg, J.~Vanderplas, A.~Passos,
  D.~Cournapeau, M.~Brucher, M.~Perrot, and E.~Duchesnay.
\newblock Scikit-learn: Machine learning in {P}ython.
\newblock {\em Journal of Machine Learning Research}, 12:2825--2830, 2011.

\bibitem{matplot}
J.~D. Hunter.
\newblock Matplotlib: A 2d graphics environment.
\newblock {\em Computing in Science \& Engineering}, 9(3):90--95, 2007.

\bibitem{networkX}
Aric~A. Hagberg, Daniel~A. Schult, and Pieter~J. Swart.
\newblock Exploring network structure, dynamics, and function using networkx.
\newblock In Ga\"el Varoquaux, Travis Vaught, and Jarrod Millman, editors, {\em
  Proceedings of the 7th Python in Science Conference}, pages 11 -- 15,
  Pasadena, CA USA, 2008.

\end{thebibliography}

\end{document}